\documentclass[11pt,a4paper]{article}
\usepackage[hyperref]{acl2019}
\usepackage{times}
\usepackage{latexsym}
\usepackage[colorinlistoftodos,prependcaption,textsize=footnotesize]{todonotes}

\usepackage{graphicx}
\usepackage{caption}

\usepackage{amsmath}
\usepackage{multirow}
\usepackage{booktabs}
\usepackage{float}
\usepackage{amssymb}

\usepackage{url}

\usepackage{amsmath,amssymb,amsthm}
\usepackage{subcaption,siunitx,booktabs}

\aclfinalcopy 

\title{The Unbearable Weight of Generating Artificial Errors for Grammatical Error Correction}

\author{Phu Mon Htut\thanks{\:Work done during internship at Grammarly} \\
Center for Data Science \\
  New York University \\
  {\tt pmh330@nyu.edu} \\\And
  Joel Tetreault \\
  Grammarly \\
  {\tt  joel.tetreault@grammarly.com} \\}
  
\date{}

\begin{document}
\maketitle
\begin{abstract}
 In recent years, sequence-to-sequence models have been very effective for end-to-end grammatical error correction (GEC).
As creating human-annotated parallel corpus for GEC is expensive and time-consuming, there has been work on artificial corpus generation with the aim of creating sentences that contain realistic grammatical errors from grammatically correct sentences. In this paper, we investigate the impact of using recent neural models for generating errors to help neural models to correct errors.  We conduct a battery of experiments on the effect of data size, models, and comparison with a rule-based approach.
 
 
\end{abstract}

\section{Introduction}
%



Grammatical error correction (GEC) is the task of automatically identifying and correcting the grammatical errors in the written text. 
Recent work treats GEC as a translation task that use sequence-to-sequence models \citep{SutskeverVL14seq2seq,BahdanauCB14} to rewrite sentences with grammatical errors to grammatically correct sentences. As with machine translation models, GEC models benefit largely from the amount of parallel training data. Since it is expensive and time-consuming to create annotated parallel corpus for training, there is research into generating sentences with artificial errors from grammatically correct sentences with the goal of simulating human-annotated data in a cost-effective way \citep{YuanB16,XieAAJN16,chollampatt2018mlconv}. 


Recent work in artificial error generation (AEG) is inspired by the back-translation approach of machine translation systems \citep{DBLP:conf/acl/SennrichHB16,DBLP:journals/corr/abs-1804-06189}.  In this framework, an intermediate model is trained to translate correct sentences into errorful sentences.  A new parallel corpus is created using the largely available grammatically correct sentences and the corresponding synthetic data generated by this intermediate model. The newly created corpus with the artificial errors is then used to train a GEC model \citep{DBLP:conf/bea/ReiFYB17,DBLP:conf/naacl/XieGXNJ18,Ge2018HumanGEC}.

To date, there is no work that compares how different base model architectures perform in the AEG task. In this paper, we investigate how effective are 
different model architectures in generating artificial, parallel data to improve a GEC model.  Specifically, we train four recent neural models (and one rule-based model \cite{Bryant2018LanguageMB}), including two new syntax-based models, for generating as well as correcting errors.   We analyze which models are effective in the AEG and correction conditions as well as by data size.  Essentially, we seek to understand how effective are recent sequence-to-sequence (seq2seq) neural model as AEG mechanisms ``out of the box."



\section{Related Work}



Before the adoption of neural models, early approaches to AEG involved identifying error statistics and patterns in the corpus and applying them to grammatically correct sentences \citep{BrockettDG06,RozovskayaR10}. Inspired by the back-translation approach, recent AEG approaches inject errors into grammatically correct input sentences by adopting methods from neural machine translation  \citep{FeliceY14,KasewaS018}. \citet{DBLP:conf/naacl/XieGXNJ18} propose an approach that adds noise to the beam-search phase of an back-translation based AEG model to generate more diverse errors. They use the synthesized parallel data generated by this method to train a multi-layer convolutional GEC model and achieve a 5 point $F_{0.5}$ improvement on the CoNLL-2014 test data \cite{ng-EtAl:2014:W14-17}. \citet{Ge2018HumanGEC} propose a fluency-boosting learning method that generates less fluent sentences from correct sentences and pairs them with correct sentences to create new error-correct sentence pairs during training. Their GEC model trained with artificial errors approaches human-level performance on multiple test sets.

\section{Approach}

\subsection{Correction and Generation Tasks}
We train our models on the two tasks---error correction and error generation. In \textit{error correction}, the encoder of the sequence-to-sequence model takes an errorful sentence as  input and the decoder outputs the grammatically correct sentence. The process is reversed in the \textit{error generation} task, where the model takes a correct sentence as input and produces an errorful sentence as the output of the decoder. 

We investigate four recent neural sequence-to-sequence models---(i) multi-layer convolutional model \citep[MLCONV;][]{chollampatt2018mlconv}, (ii) Transformer \citep{DBLP:conf/nips/VaswaniSPUJGKP17}, (iii) Parsing-Reading-Predict Networks \citep[PRPN;][]{shen2018neural}, (iv) Ordered Neurons \citep[ON-LSTM;][]{shen2018ordered}---as error correction models as well as error generation models. The PRPN and ON-LSTM models are originally designed as recurrent language models that jointly learn to induce latent constituency parse trees. We use the adaption of PRPN and ON-LSTM models as decoders of machine translation systems \citep{anon2018syntaxtranslation}: In this setting, a 2-layer LSTM is used as the encoder of the syntactic seq-to-seq models, and the PRPN and ON-LSTM are implemented as the decoders with attention \citep{BahdanauCB14}. We hypothesize that syntax is important in GEC and explore whether models that incorporate syntactic bias would help with GEC task. We provide a brief description of each model in \S\ref{sec:models} and refer readers to the original work for more details. 

\subsection{Models}
\label{sec:models}
\paragraph{Multi-layer Convolutional Model}
We use the multi-layer convolutional encoder-decoder base model (MLCONV) of \citet{chollampatt2018mlconv} using the publicly available code from the authors.\footnote{\url{https://github.com/nusnlp/mlconvgec2018}} 
As our aim is to only compare the performance of different architectures and not to achieve state-of-the-art performance, we make few changes to their code. 
The model of \citet{chollampatt2018mlconv} produces 12 possible correct sentences for each input sentences with error. They also train an N-gram language model as a re-ranker to score the generated sentences and pick the corrected sentence with the best score as final output. We did not use this re-ranking step in our model, nor did we perform ensembling or use the pre-trained embeddings as in the original work. We do not observe improvement in models like transformer and PRPN using re-ranking with an N-gram language model. Additionally, there's only a slight improvement in MLCONV using re-ranking. The reason might be because the N-gram language model is not very powerful.

\paragraph{Transformer Model} We use the publicly available Fairseq framework which is built using Pytorch for training the Transformer model. We apply the same hyper-parameters used for training the IWSLT'14 German-English translation model in the experiments of \citet{DBLP:conf/nips/VaswaniSPUJGKP17}.

\paragraph{PRPN Model} is a language model that jointly learns to parse and perform language modeling \citep{shen2018neural}. It uses a recurrent module with a self-attention gating mechanism and the gate values are used to construct the constituency tree. 
We use the BiLSTM model as the encoder and PRPN as the decoder of the sequence-to-sequence model. 

\paragraph{ON-LSTM Model}
 is follow-up work of PRPN, which incorporates syntax-based inductive bias to the LSTM unit by imposing hierarchical update order on the hidden state neurons \citep{shen2018ordered}. ON-LSTM assumes that different nodes of a constituency trees  are represented by the different chunks of adjacent neurons in the hidden state, and introduces a master forget gate and a master input gate to dynamically allocate the chunks of hidden state neurons to different nodes. 
 We use a BiLSTM model as encoder and ON-LSTM model as decoder.


\section{Experiments}

\subsection{Data}
We use the NUS Corpus of Learner English \citep[NUCLE;][]{DBLP:conf/bea/DahlmeierNW13} and the Cambridge Learner Corpus \citep[CLC;][]{Nicholls2003} as base data for training both the correction and generation models.  We remove sentence pairs that do not contain errors during preprocessing resulting in 51,693 sentence pairs from NUCLE and 1.09 million sentence pairs from the CLC .
We append the CLC data to the NUCLE training set (henceforth NUCLE-CLC) to use as training data for both AEG and correction. We use the standard NUCLE development data as our validation set and we early-stop the training based on the cross-entropy loss of the seq-to-seq models for all models.  For the generation of synthetic errorful data, we use the 2017 subsection of the LDC New York Times corpus also employed in the error generation experiments of \citet{DBLP:conf/naacl/XieGXNJ18} which contains around 1 million sentences.\footnote{\url{https://catalog.ldc.upenn.edu/LDC2008T19} }

\begin{table*}[]
\small
\centering
\begin{center}
\begin{tabular}{llccccccc} 
\toprule
 \bf GEC Model & \bf AEG model & \bf NUCLE-CLC &  \bf 10K  & \bf  50K & \bf 100K  & \bf  500K & \bf 1M & \bf 2M \\
 \midrule
 MLCONV  & MLCONV  & 35.2 & 35.1 &  34.7 &  34.6 &  38.9 & 39.4 & 34.0   \\
 Transformer & MLCONV  & 36.3 &  43.9 & \bf 44.1   & \bf 45.4 &  \bf 44.4 & \bf 45.5 &  \bf 42.0  \\
 PRPN & MLCONV  & \bf 43.6 &  \bf 45.4  &  42.8 &  43.2 &  39.6 &  38.6  & 31.7 \\
 ON-LSTM & MLCONV  & 36.6  &  39.8   & 35.6 & 38.4  &  36.9 &  24.2  & 20.1 \\
 \midrule
 MLCONV  & Transformer & 35.2 & 36.1 &  35.2 &   39.4 & 36.6 & 36.6 & 36.1   \\
 Transformer & Transformer & 36.3 & 20.1 & \bf 43.9   & \bf 42.9 & \bf 43.7 &  \bf  44.0 & \bf 41.0  \\
 PRPN & Transformer &  \bf 43.6 & \bf 43.1  &  40.9 &  40.6 & 41.4 &  29.4  & 31.7 \\
 ON-LSTM & Transformer & 36.6  &  39.8   & 38.2 & 39.6  & 24.0 &  21.3  & 20.1 \\
  \midrule
 MLCONV & Rule-based &  35.2 & 6.0 &  7.8 &  10.5 &  13.7 & 13.9 & --  \\
 Transformer  & Rule-based &  36.3 & \bf 13.5 &  \bf 14.4   &  \bf 21.8 & \bf 14.5 & \bf 21.6 & --  \\
 PRPN  & Rule-based & \bf  43.6 & 2.8  &  4.9 &  2.6 &  3.9 &  8.9 & --  \\
 ON-LSTM  & Rule-based &  36.6  & 4.7   & 3.9 & 5.5  &  4.2 &  5.3 & --  \\
  \bottomrule  
\end{tabular}
\end{center}
\caption{\label{tab:rb-table} (Exp2) Evaluating the impact of MLCONV, Transformer and the rule-based AEG systems.  
NUCLE-CLC column represents the F0.5 score of GEC models trained on the base NUCLE-CLC data. \textit{10K, 50K, 100K, 500K, 1M, and 2M} represents the amount of artificial data added to the NUCLE-CLC during training. 
}
\end{table*}

\subsection{Setup}
We conduct four experiments in this paper.  In \textbf{Exp1}, we train all the AEG models and intermediate GEC models on NUCLE-CLC. We use the NYT dataset as input to the AEG models to generate sentences with artificial errors.  We then create new parallel training sets for correction by combining the sentences from CLC and NUCLE with the errorful sentences generated by each model. We then train the GEC models using these parallel datasets. 

The three other experiments are variants of the first. In \textbf{Exp2} we train all correction models on artificial errors generated by the top neural AEG systems and a rule-based system for comparison.  In \textbf{Exp3}, we train the GEC models on NUCLE to analyze models built on real data.   Finally, in \textbf{Exp4}, we train all GEC models on artificial data to determine how well correction models can perform without any real data.


All our experiments are tested on the CoNLL-2014 test set and we use the sentence-level $F0.5$ score from the MaxMatch ($M^2$) scorer \citep{DahlmeierN12} for evaluation.  All models are implemented using the Fairseq framework.\footnote{\url{https://github.com/pytorch/fairseq}}


\subsection{Results}

\begin{figure}[htbp]
    \centering
        \includegraphics[height=1.8in] {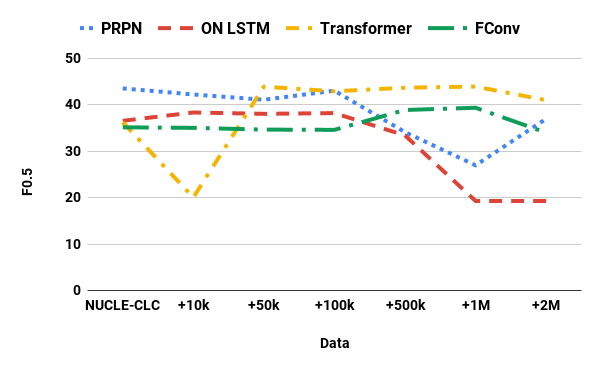}
        \caption{(Exp1) Models trained on the artificial data generated by the corresponding AEG model.  The X-axis represents the amount of artificial data added to NUCLE-CLC during training. 
        }
        \label{fig:table1a}
\end{figure}

\noindent\textbf{Exp1}: Figure~\ref{fig:table1a} shows the performance of GEC models trained on the base NUCLE-CLC set and then retraining with various amounts of artificial data.  We first observe that PRPN performs substantially higher than the rest of the models when trained only with the base CLC-NUCLE data.  However, its performance drops when artificial data generated by the corresponding PRPN AEG model is added. As for ON-LSTM, the performance improves slightly when the amount of added data is less than 100k but the performance drops drastically otherwise. Conversely, the performance of MLCONV and Transformer improves with the added artificial data but the improvement is not linear with the amount of added data. 

\noindent\textbf{Exp2}: Since the performance of MLCONV and Transformer GEC models improve with the addition of artificial data generated by corresponding AEG models, we hypothesize that the artificial error generated by these models are useful. To test this hypothesis, we train all the GEC models with various amount of artificial error generated by MLCONV and Transformer AEG models.  We also compare these AEG models to a rule-based one inspired by the confusion set generation method in \citet{Bryant2018LanguageMB}. We subsequently score each sentence with a language model (GPT-2 \citep{noauthororeditor}) in order not to select the most probable sentence. This method generates a confusion set for prepositions (set of prepositions plus an empty element), determiners, and morphological alternatives (cat $\rightarrow$ cats).

The results of these experiments are found in Table~\ref{tab:rb-table}.  Nearly all correction models improve when using MLCONV or Transformer AEG data with the biggest performances yielded using the Transformer model.  Interestingly, when using 1M or 2M samples, performance starts to decline.  We believe that over 1M samples, the noisiness of the artificial data overwhelms the contributions of the real data (roughly over 1M samples).  The performance of all models drops when trained with the errors generated by the rule-based model. It is interesting to observe that the performance drops significantly just by adding 10K artificial sentences to the base data.

\noindent\textbf{Exp3}: Table \ref{tab:nucle-table} shows the performance of the models trained on NUCLE dataset with additional artificial data generated by corresponding AEG models trained on NUCLE-CLC. We can see that the performance of all models, except ON-LSTM, improves significantly when 1 million artificial sentence pairs are added to the NUCLE training data, even though the errors in these sentences do not resemble natural errors. This contrasts with the result in Figure~\ref{fig:table1a} where the performance of the GEC models trained with the combination of artificial error and CLC-NUCLE base data drops. This suggests that artificial data is helpful when the base data size is relatively small.

\begin{table}[htbp]
\small
\centering
\setlength{\tabcolsep}{9 pt} 
\begin{center}
\begin{tabular}{lcccc} 
\toprule
 \bf Model & \bf NUCLE &  \bf +10K  & \bf  +50K &  \bf +1M \\
 \midrule
 MLCONV  & 10.1 & 12.3 &  12.9 &   16.1   \\
 Transformer & 11.2   & \bf 28.1 &  \bf 16.9  &   22.8  \\
 PRPN  & 8.3 & 6.9 &  12.5 &  \bf 26.2   \\
 ON-LSTM & 9.4   & 11.3 &  11.8  &  6.0  \\
  \bottomrule  
\end{tabular}
\end{center}
\caption{\label{tab:nucle-table} (Exp3) Using only NUCLE as base training for correction.  The AEG models are trained using NUCLE-CLC data as in other experiments. 
}
\end{table}

\noindent\textbf{Exp4}: The GEC models trained only on artificial data perform very poorly. The best setups, Transformer and MLCONV, achieve F0.5 scores of 12.8 and 12.4 respectively when trained with 2 million sentences generated by the corresponding AEG model. 
This outcomes suggests that AEG data should be paired with some sample of real data to be effective.

\subsection{Manual Evaluation}
We performed a manual analysis of the generated error sentences and found that many of the errors did not always resemble those produced by humans.  For example, \textit{The situation with other types is not much (better $\rightarrow$ downward)}.  This shows that despite the noisiness of the error-generated data, some models (namely MLCONV and Transformer) were robust enough to improve. This also suggests that we may achieve better improvement by controlling artificial errors to resemble the errors produced by humans. The performance of syntax-based models goes down significantly with the addition of artificial errors, which indicates that these models may be sensitive to poor artificial data.



%


\section{Conclusion}
We investigated the potential of recent neural architectures, as well as rule-based one, to generate parallel data to improve neural GEC.  We found that the Multi-Layer Convolutional and Transformer models tended to produce data that could improve several models, though too much of it would begin to dampen performance.  The most substantial improvements could be found when the size of the real data for training was quite small.  We also found that the syntax-based models, PRPN and ONLSTM, are very sensitive to the quality of artificial errors and their performance drops substantially with the addition of artificial error data.  Our experiments suggest that, unlike in machine translation, it is not very straightforward to use a simple back-translation approach for GEC as unrealistic errors produced by back-translation can hurt the correction performance substantially.

We believe this work shows the promise of using recent neural methods in an out-of-the-box framework, though with care.  Future work will focus on ways of improving the quality of the synthetic data.  Ideas include leveraging recent developments in powerful language models or better controlling for diversity and frequency of specific error types.

\section*{Acknowledgements}
We would like to thank the Grammarly Research Team, especially Maria Nadejde, Courtney Napoles, Dimitris Alikaniotis, Andrey Gryschuck, Maksym Bezva and Oleksiy Syvokon.  We would also like to thank Sam Bowman, Kyunghyun Cho, and the three anonymous reviewers for their helpful discussion and feedback.




\bibliography{acl2019}
\bibliographystyle{acl_natbib}

\end{document}